\documentclass[letterpaper]{article} 
\usepackage{aaai25}  
\usepackage{times}  
\usepackage{helvet}  
\usepackage{courier}  
\usepackage[hyphens]{url}  
\usepackage{graphicx} 
\usepackage{natbib}  
\usepackage{caption} 
\usepackage{algorithm}
\usepackage{algorithmic}

\usepackage{newfloat}
\usepackage{listings}
\DeclareCaptionStyle{ruled}{labelfont=normalfont,labelsep=colon,strut=off} 
\floatstyle{ruled}
\newfloat{listing}{tb}{lst}{}
\floatname{listing}{Listing}
\usepackage{multirow}
\usepackage{booktabs}
\usepackage{amsmath}
\usepackage{amssymb}
\usepackage{hyperref}

\title{Population Aware Diffusion for Time Series Generation}
\author {
    Yang Li\textsuperscript{\rm 1,},
    Han Meng\textsuperscript{\rm 1},
    Zhenyu Bi\textsuperscript{\rm 2},
    Ingolv T. Urnes \textsuperscript{\rm 3},
    Haipeng Chen \textsuperscript{\rm 1}
}
\affiliations {
    \textsuperscript{\rm 1}William \& Mary \\
    \textsuperscript{\rm 2}Virginia Tech \\
    \textsuperscript{\rm 3}Generated Health\\
    \textsuperscript{\rm 1} \{yli102, hmeng, 
 hchen23\}@wm.edu, \textsuperscript{\rm 2} zhenyub@vt.edu,\textsuperscript{\rm 3} ingolv.urnes@generatedhealth.com
    
}

\usepackage{bibentry}

\begin{document}

\maketitle

\begin{abstract}
Diffusion models have shown promising ability in generating high-quality time series (TS) data. Despite the initial success, existing works mostly focus on the authenticity of data at the \textit{individual} level, but pay less attention to preserving the \textit{population}-level properties on the entire dataset. Such population-level properties include value distributions for each dimension and distributions of certain functional dependencies (e.g., cross-correlation, CC) between different dimensions. 
For instance, when generating house energy consumption TS data, the value distributions of the outside temperature and the kitchen temperature should be preserved, as well as the distribution of CC between them. 
Preserving such TS population-level properties is critical in maintaining the statistical insights of the datasets, mitigating model bias, and augmenting downstream tasks like TS prediction.   
Yet, it is often overlooked by existing models. Hence, data generated by existing models often bear distribution shifts from the original data. 
We propose {\bf P}opulation-{\bf a}ware {\bf D}iffusion for {\bf T}ime {\bf S}eries {\bf(PaD-TS)}, a new TS generation model that better preserves the population-level properties. The key novelties of PaD-TS include 1) a new training method explicitly incorporating TS population-level property preservation, and 2) a new dual-channel encoder model architecture that better captures the TS data structure. Empirical results in major benchmark datasets show that PaD-TS can improve the average CC distribution shift score between real and synthetic data by 5.9x while maintaining a performance comparable to state-of-the-art models on individual-level authenticity. 
\end{abstract}

\begin{links}
    \link{Code}{https://github.com/wmd3i/PaD-TS}
\end{links}

\section{Introduction}

Time series data exists in a broad spectrum of real-world domains, spanning healthcare \cite{kaushik2020ai,morid2023time}, energy \cite{priesmann2021time,deb2017review}, finance \cite{zeng2023financial,masini2023machine}, and many more.
TS models have been used in these domains for effective data analysis and prediction tasks. Developing such models requires rich and high-quality TS datasets, which unfortunately may not exist in many data-scarce domains like healthcare and energy. Various data augmentation techniques (e.g., jittering, scaling, permutation, time warping, and window slicing) have been developed to enhance the original datasets with synthetically generated TS data.
Synthetic data can potentially create observations that do not exist but are close to the original dataset \cite{coletta2023constrained,esteban2017real}. With a newly augmented dataset, we can further enhance TS models for data analysis, while protecting the privacy and confidentiality of the original data and potentially enhancing data sharing. 

\begin{figure}[t]
\centering
\includegraphics[width=1\columnwidth]{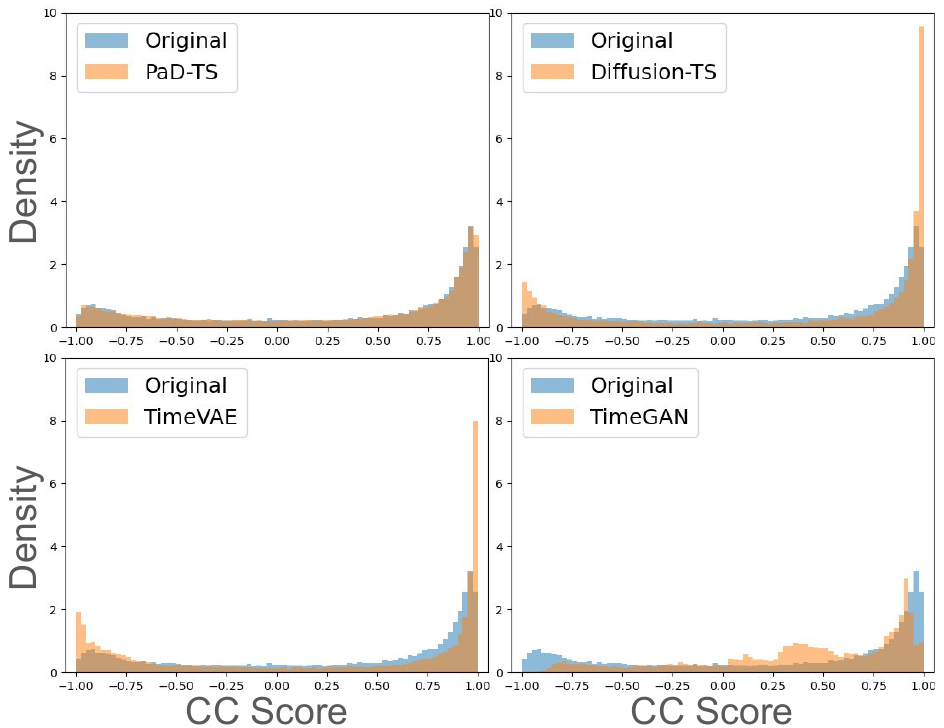} 
\caption{Histogram of CC distribution between the original and synthetic Energy datasets. The CC values are calculated between the outside temperature and kitchen temperature. 
PaD-TS (top left) best preserves such functional dependency distribution. Previous models tend to generate data points with a CC score close to 1 or -1, which leads to biases for downstream tasks.}
\label{Fig:CC_dist}
\end{figure}

How do we measure the quality of synthetic TS data? First, it should be authentic on the \textit{individual} level. Given two samples -- one from the original set and another from the generated set -- we should not be able to identify which is fake or real. Second, it should preserve the TS \textit{population}-level properties of the original data. Such population-level properties include distributions for each dimension of the data and distributions of certain functional dependencies (e.g., CC) between different dimensions of the data. 
Taking house energy consumption TS as an example \cite{candanedo2017data}, the value distribution of the outside temperature and the kitchen temperature, as well as the CC distribution between them should be preserved. As shown in Figure \ref{Fig:CC_dist},
previous methods trust the model to estimate the CC distribution between them which fails to preserve the same distribution in generated TS.
Preserving both the individual- and population-level properties is crucial in maintaining the standalone and statistical insights of the original data, hence reducing model bias and further augmenting downstream tasks such as prediction.

We revisit the TS generation problem with an emphasis on preserving the TS population-level properties. There have been extensive studies on TS generation using generative adversarial networks (GANs) \cite{yoon2019time,liao2020conditional,CRNNGAN,esteban2017real,pei2021towardsGAN} and variational autoencoders (VAEs) \cite{desai2021timevae,naiman2023generative}. Though they have achieved reasonable results, it is known that GANs suffer from unstable training because of the need to interactively and iteratively train both the generator and discriminator, while VAEs normally generate lower-quality samples due to optimizing an approximate objective via the evidence lower bound (ELBO). Moreover, both GANs and VAEs may struggle with the mode collapse issue. Diffusion models (DMs) emerge as another class of powerful generative models that are robust against mode collapse, and show state-of-the-art performances in domains such as image generation \cite{ho2020denoising, nichol2021improved,peebles2023scalable,sohl2015deep}, text-to-image generation \cite{ramesh2022hierarchical,nichol2021glide}, and text-to-video generation \cite{videoworldsimulators2024}. In light of this, recent studies have developed DM-based TS generation models \cite{coletta2023constrained,yuan2024diffusionts} that yield better results than GANs and VAEs for TS generation. Despite the initial success, most existing works pay less attention to the preservation of population-level properties and hence may suffer from the generation distribution shift.

We hypothesize that the distribution shift of existing TS generative models comes from two sources: 1) TS population-level property preservation is not explicitly incorporated into the training process, as they only try to capture TS population-level properties by minimizing the value distance between synthetic and original samples. 
InfoVAE \cite{zhao2017infovae} tackles a similar issue in image generation by penalizing the distribution shift in the single encoded latent space with a regularization term in its training loss. 
However, this cannot be directly applied to DMs, as DMs follow an iterative generation framework usually with an extremely long series of latent spaces.
2) Model architectures of existing works cannot fully capture the multivariate TS data information. Cross-dimensional information has been shown to be critical \cite{liu2024itransformer} for TS prediction, as it can yield great performance using only dimension information.
Previous methods either neglect cross-dimension features \cite{yuan2024diffusionts,yoon2019time,desai2021timevae} or try to capture them in a shared block with temporal feature extractions
\cite{coletta2023constrained,tashiro2021csdi}, which may not be sufficient to capture all cross-dimension features.  

We propose PaD-TS, a new DM that addresses the above issues. 
PaD-TS comes with a new DM training objective that penalizes population-level distribution shifts. This is enabled by a new sampling strategy (during training) that enforces the comparison of two distributions to the same diffusion step in a mini-batch. 
In addition, we design a new transformer \cite{peebles2023scalable,vaswani2017attention} encoder-based dual-channel architecture that can better capture the population-level properties.

Our main contributions are as follows. 
1) We are the first to study DM-based TS generation that explicitly considers TS population-level property preservation, along with new metrics to evaluate it. 
2) We propose PaD-TS, a novel DM that addresses the technical challenges of this problem. 
3) We conduct extensive empirical evaluations of our model. The empirical results show that PaD-TS achieves state-of-the-art performance in population-level property preservation and comparable individual-level authenticity.

\section{Related Work} \label{relatedwork}
\textbf{GANs} \cite{goodfellow2014generative} are 
generative models that usually consist of a generator and a discriminator. The generator generates plausible data to fool the discriminator, whereas the discriminator tries to distinguish the synthetic data from real data. Due to their ability to generate high-quality synthetic data \cite{CRNNGAN,yoon2019time,liao2020conditional}, GANs have been extensively studied for TS generation \cite{CRNNGAN,yoon2019time,liao2020conditional} as well as functional dependency persevation tables \cite{chen2019faketables}.
However, GANs suffer from inherent instability, resulting from the interactive and iterative training process on both the generator and the discriminator. This often leads to non-converging models that oscillate or vanishing gradient issues that prevent the generator from learning meaningful patterns.

\textbf{VAEs} \cite{kingma2022autoencodingvariationalbayes} are another family of generative models based on the encoder-decoder architecture. The encoder in VAEs encodes the data in a latent space following a Gaussian distribution. The decoder samples from learned latent space as prior information to generate synthetic data.
TimeVAE \cite{desai2021timevae} utilizes convolution structures to capture temporal correlations. KoVAE \cite{naiman2023generative} uses a linear Koopman-based prior and a sequential posterior to further improve the performance.
However, VAEs are known for their low generation quality and susceptibility to mode collapse, which limits their success in TS generation.

\textbf{DMs} \cite{ho2020denoising,sohl2015deep} emerge as a new generative framework that learns to generate data by gradually reversing a noising process applied to the training data.
Being theoretically grounded with connections to score-based generative modeling \cite{song2020score}, and having robustness against mode collapse compared to other generative models like GANs and VAEs, they are arguably the state-of-the-art methods in image generation \cite{ho2020denoising, nichol2021improved,peebles2023scalable,sohl2015deep}, text-to-image generation \cite{ramesh2022hierarchical,nichol2021glide}, and text-to-video generation \cite{videoworldsimulators2024}. 
Recent studies have developed DM-based models for TS generation \cite{coletta2023constrained,yuan2024diffusionts} that outperform GAN- and VAE-based models, further demonstrating the superior performance of DMs.
Diffusion-TS \cite{yuan2024diffusionts} uses trends, seasonal architectures, and Fourier-based objects, which outperform previous models by a significant margin. TimeLDM \cite{qian2024timeldmlatentdiffusionmodel} combines VAE and the diffusion framework. 
In addition to TS generation, DMs have been widely used for TS prediction \cite{feng2024latent,li2022generative,rasul2021autoregressive,alcaraz2022diffusion,yang2024generative} and imputation \cite{tashiro2021csdi}, extremely long TS generation \cite{barancikova2024sigdiffusions}, and general pre-trained models for TS prediction \cite{yang2024generative}. These works do not explicitly consider the preservation of population-level properties and hence may suffer from the generation shift, a key issue that we are going to address in this paper.

\section{Problem Statement} \label{PS}

Given a multivariate TS dataset $\mathcal{D}_{\text{orig}}=\{x_n\}_{n=1}^N$ with $N$ samples. Each data sample $x^{1:L; 1:F}  \in \mathbb{R}^{L \times F}$ is a multivariate TS, where $L$ is the sequence length and $F$ is the number of features/dimensions (e.g., we can denote $\{x^{l;i}\}$ for all $l \in [1,L] $ as values in the $l$-th dimension). 
Our task is to generate synthetic dataset $\mathcal{D}_{\text{syn}}=\{\hat{x}_n\}$ such that the synthetic data is similar to the original data $\mathcal{D}_{\text{orig}}$ in individual level and follows original population-level property distributions.
To evaluate the generation quality at the individual level, we have the following $metric$:

(1) \textit{Discriminative Accuracy (DA)} \cite{yoon2019time} is based on the post-hoc machine learning classifier (clf) trained with the training set from original and synthetic datasets (with label real=1; synthetic = 0). Then DA is the model performance on the test set with size $S$ using the following equation: 
\begin{equation}
    \text{DA}\!=\bigr|\frac{\sum_{n=1}^S (0 \!=\! \text{clf}(\hat{x}_n)) \!+\! \sum_{n=1}^S (1 \!=\! \text{clf}(x_n))}{2S} \!-\! 0.5 \bigr|
\end{equation}
where $\hat{x}_n$ and $x_n$ are test samples.  
To evaluate the generation quality in terms of TS population-level property preservation, we propose two new \textit{metrics}: 

(2) \textit{Value distribution shift (VDS)}: 
\begin{equation}
    \text{VDS} = \frac{1}{F}\sum_{i =1 }^F D(P_V^i, Q_V^i)
\end{equation}
where $D$ stands for a certain distribution distance measure (e.g., KL divergence); $P_V^i$ is the value distribution of $i$-th dimension over the original data 
; and $Q_V^i$ is the counterpart distribution for the synthetic dataset.

(3) \textit{Functional dependency distribution shift (FDDS)}: 
\begin{equation}
    \text{FDDS}= \frac{1}{M}\sum_{m = 1}^{M} D(P^{i,j}_{\text{FD}}, Q^{i,j}_{\text{FD}})
\end{equation}
where $P_{\text{FD}}^{i,j}$ is the distribution of the functional dependency scores between $i$-th and $j$-th dimension over the original data which can be calculated by any functional dependency function of interest $f: x^{1:L; i}\times x^{1:L; j} \rightarrow \mathbb{R}$ (e.g., cross-correlation $CC$, mutual information, etc.); $Q_{\text{FD}}$ is the counterpart distribution for the synthetic dataset; and $M$ represents the possible pairs of functional dependencies.

New metrics (VDS and FDDS) with reasonable distribution distance measure a more general similarity between real and synthetic population-level property distributions. They yield better performance in detecting potential biases and shifts in generated samples than aggregated statistics-based metrics (e.g., distance between property mean).

\section{Approach}

In this section, we introduce PaD-TS which addresses population-level preservation problems. Building on top of diffusion models, PaD-TS consists of two novel components: a new population-aware training process, and a new dual-channel encoder model architecture. 

\subsection{Preliminary: Diffusion Models} \label{ddpm}
We briefly review the formulations of the denoising diffusion probabilistic models (DDPMs) \cite{ho2020denoising,nichol2021improved} which have iterative \textit{forward} and \textit{reverse} processes.

Given original data $x^0 \sim q(x^0)$, the \textit{forward} process $q$ is a Markov process which adds noise $\epsilon^t$ iteratively based on a fixed variance scheduler $\beta^t$ and sampled diffusion step $t$. Normally, $t$ is uniformly sampled from $[1,T-1] $ for each sample data.
\begin{equation} \label{Eq:4}
\begin{aligned}
    q(x^{1:T}|x^0) &= \prod_{t=1}^T q(x^t|x^{t-1}) \\ 
     q(x^t| x^{t-1}) &= \mathcal{N} (\sqrt{1-\beta^t} x^{t-1}, \beta^t \textbf{I})
\end{aligned}
\end{equation}
The \textit{reverse} process ($p_{\theta})$ gradually removes noises from $p(x^T) \sim \mathcal{N} (0,\textbf{I})$ and tries to recover $x^0$ with $x^0(\theta)$. 
\begin{equation}
\begin{aligned}
    &p_\theta(x^{0:T}) = p(x^T)\prod_{t=1}^T p_{\theta}(x^{t-1}|x^{t}) \\ 
    &p_{\theta}(x^{t-1}| x^{t}) = \mathcal{N} (\mu_{\theta}, \Sigma_{\theta})
\end{aligned}
\end{equation}
where mean $\mu_{\theta}$ and variance $\Sigma_{\theta}$ are learnable parameters.

To reduce complexity, DDPMs set $\Sigma_{\theta} = \sigma_t^2\textbf{I}$ where $\sigma_t^2 = \beta^t$ follows the same variance scheduler as the forward process. With fixed $\Sigma_{\theta}$, one can effectively approximate the reverse process by training a model that predicts $\mu_{\theta}$ with: 
\begin{equation}
    \begin{aligned}
    \Tilde{\mu}^t(x^t,x^0) &= \frac{\sqrt{\bar{\alpha}^{t-1}} \beta^t}{1-\bar{\alpha}^t}x^0 + \frac{\sqrt{\alpha^t}(1-\bar{\alpha}^{t-1})}{1-\bar{\alpha}^t}x^t
    \end{aligned}
\end{equation}
where $\alpha^t = 1 - \beta^t$ and $\bar{\alpha}^t = \prod_{s=1}^{t}\alpha_s$ are both constants.

Note that the only unknown in $\Tilde{\mu}^t (x^t,x^0)$ is $x^0$. A neural network can directly model towards $x^0$ or $\epsilon^t$ when applying reparametrization trick $x^0 = \frac{1}{\sqrt{\bar{\alpha}^t}}(x^t - \sqrt{1-\bar{\alpha}^t}\epsilon^t)$. 
With target $x^0$, DDPMs have the following training objective:
\begin{equation}
    \label{l0}
    L_0(\theta) = \mathbb{E}_{t,x^0}\Bigr[\| x^0 - x^0(\theta)\|^2\Bigr]
\end{equation}

\begin{algorithm}
\caption{PaD-TS training procedure}
\label{alg:algorithm}
\textbf{Input}: Original TS data, FD function $f$, epochs $E$, and total diffusion steps $T$\\
\textbf{Output}: Trained PaD-TS model $\theta$
\begin{algorithmic}[1] 
\FOR{$i=1$ to $E$} 
    \STATE Sample a mini-batch of $x^0$ with $b$ samples
    \STATE Sample $t_1 \in [1,T-1]$ \hfill $\rhd\!\!\rhd\!\!\rhd$ SSS
    \STATE Let $t = [t_1,...,t_1]$ 
    \STATE Get $\hat{x}^0$ using PaD-TS model \hfill $\rhd\!\!\rhd\!\!\rhd$ PAT objective
    \STATE Find all FD distributions for $\hat{x}^0$ and $x^0$ 
    \STATE Calculate $L_0$ and $L_{pop}$ 
    \STATE Update $\theta$ with gradient $\nabla_\theta (L_0 + L_{pop})$  
\ENDFOR
\STATE \textbf{return} Model $\theta$
\end{algorithmic}
\end{algorithm}
\subsection{PaD-TS Training}
As mentioned above, existing DMs show promising performance in individual-level authenticity but exhibit suboptimal performance in preserving the TS distribution of population-level properties. One hypothesis is that the original DDPM training process focuses on the value distance between model input and output, and overlooks TS population-level properties preservation.
A naive resolution to this is to penalize distribution shifts by regularizing the loss function. However, this turns out to be not directly feasible because of the iterative DM generation process.  
To address this technical challenge, we propose a new DM training process that enables applying any regularization of interest for DMs which consists of two components: population aware training (PAT) objectives and same diffusion step samplings (SSS).

Algorithm \ref{alg:algorithm} shows our new DM training procedure for PaD-TS: (1) We first sample a mini-batch of data from the original TS dataset in Line 2. (2) In Lines 3-4, we use the SSS for the mini-batch diffusion step $t$. (3) In Lines 5-8, we use the PAT objective to update the model parameter $\theta$. The details of the training procedure are as follows.

\subsubsection{Population Aware Training Objective}
This objective considers preserving the TS population-level property distribution rather than simple statistical measures (e.g., mean, variance, etc). Thus it is crucial to use the right distribution distance. 
The distribution of property at the population level may come in arbitrary parametric forms, thus distribution shift measures such as Kullback-Leibler divergence \cite{zhao2017infovae, liu2016stein} and the Wasserstein distance \cite{arjovsky2017wasserstein} are inappropriate, as in practice they usually either have assumptions toward the underlying distributions and/or is computationally expensive, especially with high dimensional data.
Inspired by InfoVAE \cite{zhao2017infovae}, we use the Maximum Mean Discrepancy (MMD) \cite{gretton2012kernel} as the distribution distance. It is commonly used in deep learning tasks such as image generation \cite{zhao2017infovae,li2017mmd} and domain adaptation \cite{yan2017mind,cao2018unsupervised}. 

With a pair of arbitrary distributions ($P$ and $Q$), MMD compares all their moments using the selected kernels. Using a Radial Basis Function (RBF) kernel on multiple window sizes as an example, the MMD distance can be efficiently estimated as follows:
\begin{equation} \label{MMD}
    \begin{aligned}
    &\text{MMD}_W(Q||P) =  \sum_{w_i \in W} \mathbb{E}_{Q,Q}\bigr[\text{RBF}_{w_i}(Q,Q)\bigr] + \\
    &\!\!\sum_{w_i \in W}\!\!\mathbb{E}_{Q,P}\bigr[\text{RBF}_{w_i}(Q,P)\bigr]\!\! +\! \!\!\sum_{w_i \in W}\!\mathbb{E}_{P,P}\bigr[\text{RBF}_{w_i}(P,P)\bigr]
    \end{aligned}
\end{equation}
where $\text{RBF}_{w_i}$ stands for a RBF kernel with window $w_i$.

We use cross-correlation (CC, see definition in Appendix 
\ref{CC_detail}
) as an example of TS functional dependency, as CC is often a critical property in TS data. 
For each TS data sample $x^{1:L;1:F}$, we can calculate $M = \frac{F(F-1)}{2}$ unique CC values. We use $P_{\text{CC}}^{i,j}$ to represent the distribution of CC between $i$-th and $j$-th dimension in the original data, and use $Q_{\text{CC}}^{i,j}$ to represent its counterpart for the synthetic data.
By considering all possible pairs of CC distributions, the regularization loss term can be defined as:
\begin{equation} \label{ldist}
    L_{pop} = \frac{1}{M} \sum_{m = 1}^{M} \text{MMD}_{W}(P_{\text{CC}}^{i,j}, Q_{\text{CC}}^{i,j})
\end{equation}
Hence, the PAT objective can be formally defined as follows:
\begin{equation}
    L_{total} = L_{0} + \alpha * L_{pop}
\end{equation}
where $\alpha$ is a hyperparameter that controls the weight of the population aware loss.

\begin{figure*}[t]
     \centering
     \includegraphics[width=1\linewidth]{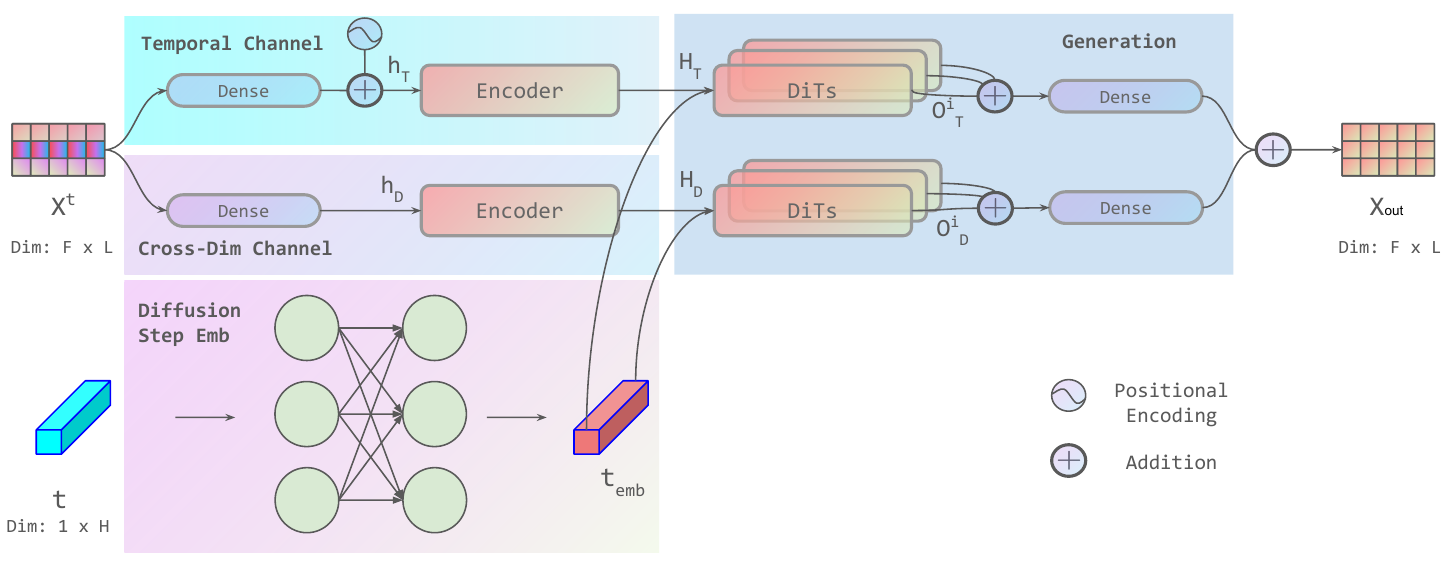}
     \caption{PaD-TS model architecture}\label{Fig:model}
\hfill
\end{figure*}

\subsubsection{Same Diffusion Step Sampling}
How do we empirically compute a meaningful $L_{pop}$ in DMs? As mentioned, the DM framework is an iterative generation process (i.e., gradually removing noise with a variance scheduler $\beta^t$) that behaves differently at each diffusion step $t$. 
A common uniform diffusion step sampling strategy \cite{yuan2024diffusionts,ho2020denoising,coletta2023constrained,peebles2023scalable} and importance sampling \cite{nichol2021improved} will produce mixed diffusion step sampling within a mini-batch. If the resulting diffusion steps of one of the two strategies are applied, DM will yield generations from mixed diffusion steps. Although it works for value distribution preservation, it will be problematic to compare functional dependency distributions because of different behaviors at different diffusion steps.
To ensure a reasonable functional dependency distribution comparison in DM, we introduce the SSS strategy.

Given a mini-batch training procedure with $b$ samples. We have diffusion step vector $t = [t_1,t_2,...,t_b]$, where each $t_i \in[0,T-1]$. SSS first samples $t_1 \in [1,T-1]$ and duplicates the diffusion step $t_1$ to fill the vector $t$. Thus we have the SSS-based diffusion step vector $t = [t_1,t_1,...,t_1]$. For a mini-batch sample, SSS ensures the distribution comparison is on the same diffusion step.   
Compared to uniform sampling,
SSS has one obvious limitation: less coverage of diffusion steps. By increasing the number of training epochs, each diffusion step in $[0,T-1]$ will eventually be sampled.

\subsection{Model Architecture}
Our model in Figure \ref{Fig:model} is based on transformer encoders including vanilla transformer \cite{vaswani2017attention} encoders and diffusion transformer (DiT) blocks \cite{peebles2023scalable}. 
To fully capture temporal and cross-dimensional information, we propose a dual-channel architecture where each part of the information is processed separately. Each channel (temporal and cross-dimensional) passes a dense layer to encode channel representation, a vanilla transformer encoder, a few residual connected DiT blocks, and a dense layer revert to its original shape.

\textbf{Temporal and cross-dimensional representation} can be learned via a linear dense layer \cite{liu2024itransformer}. Given a mini-batch TS $x \in \mathbb{R}^{b\times L \times F}$ and its corresponding diffusion steps $t$, where $ b$ stands for the number of samples in a mini-batch, L stands for sequence length, and F stands for the number of features. By permuting L and F separately, we obtain temporal first input $x_{\text{T}} \in \mathbb{R}^{b\times L \times F}$and feature first inputs $x_{\text{D}} \in \mathbb{R}^{b\times F \times L}$. For the temporal first input, we have an additional learned positional embedding pos($x_T$). This process can be formulated as follows:
\begin{equation}
    h_{T} = (W_{T} x_T + b_T) + \text{pos}(x_T)
\end{equation}
\begin{equation}
    h_{D} = W_{D} x_D + b_D
\end{equation}
where $W_T$ and $W_D$ represent dense layer parameters; $b_T$ and $b_D$ represent bias terms; and has outputs $h_T \in \mathbb{R}^{b\times L \times H}$ and $h_D \in \mathbb{R}^{b\times F \times H}$.

\textbf{Vanilla transformer encoder} (Enc) is used to analyze TS at each diffusion step.
The transformer encoder block is based on multi-head attention which is commonly used for 
pattern recognition and feature extraction \cite{peebles2023scalable, liu2024itransformer,yuan2024diffusionts, coletta2023constrained,tashiro2021csdi}. We use one transformer block for each channel to extract relative information:
\begin{equation}
    H_{T} = \text{Enc}(h_T)
\end{equation}
\begin{equation}
    H_{D} = \text{Enc}(h_D)
\end{equation}

\textbf{DiT} blocks with residual connections are the final layers in the model. Compared to the vanilla encoder blocks, \citet{peebles2023scalable} design DiTs which perform well in the diffusion framework with two advantages: high throughput and the way conditional information is introduced. Transformers are naturally with high-throughput due to the multi-head attention mechanism that can be processed in parallel. Unlike many methods that add conditional information before each block, DiTs introduce partial conditional embedding at each layer. More details can be found in Appendix \ref{app:DiT}.
In our study, conditional embedding is the diffusion step embedding $t_{emb}$ which is learned via dense layers. We use DiT blocks for the generation process which can be formally described as:
\begin{equation}
    \begin{aligned}
        O^i = &\mathbf{1}_{i= 0} \text{DiT} (H,t_{emb}) +\mathbf{1}_{i= 1} \text{DiT} (O^0 + H,t_{emb}) \\ 
        &+\mathbf{1}_{i> 0} \text{DiT} (O^{i-1} + O^{i-2},t_{emb})  
    \end{aligned}
\end{equation}
where $O^i$ is the $i$-th DiT block output, $H$ is the encoded information from the previous section, $t$ is the diffusion conditional embedding (i.e., diffusion step), and $\mathbf{1_c}$ is an indicator function with condition $c$. For the different channels, we simply replace $H$ with $H_D$ or $H_T$ to obtain $O_D^i$ or $O_T^i$.

\begin{table*}[t]
    \centering
    \begin{tabular}{llllll}
    \textbf{Metrics} & \textbf{Dataset} & \textbf{PaD-TS} & \textbf{Diffusion-TS} & \textbf{TimeGAN} & \textbf{TimeVAE}\\
    \toprule
    \multirow{3}{*}{VDS} 
    & Sines &\textbf{0.0005} &0.0007&0.0034&0.0177\\
    & Stocks&\textbf{0.0029} &0.0369&0.0257&0.0038\\
    & Energy&\textbf{0.0019}&0.0060&0.0427&0.0882\\
    \midrule
    \multirow{3}{*}{FDDS} 
    & Sines &\textbf{0.0003}&0.0031&0.0167&0.0135\\
    & Stocks&\textbf{0.0588}&0.1841&0.1117&0.2161\\
    & Energy&\textbf{0.0442}&0.1837&0.2777&0.4413\\
    \midrule
    \multirow{3}{*}{DA} 
    & Sines &0.013$\pm$ 0.004&\textbf{0.005$\pm$ 0.000}&0.037$\pm$ 0.004&0.072$\pm$ 0.061\\
    & Stocks&\textbf{0.055$\pm$ 0.087}&0.082$\pm$ 0.025&0.143$\pm$ 0.073&0.133$\pm$ 0.115\\
    & Energy&\textbf{0.078$\pm$ 0.011}&0.127$\pm$ 0.016&0.469$\pm$ 0.017&0.498$\pm$ 0.004\\
    \midrule
    \multirow{3}{*}{Predictive score} 
    & Sines &\textbf{0.093$\pm$ 0.000}&\textbf{0.093$\pm$ 0.000}&0.095$\pm$ 0.000&0.229$\pm$ 0.001\\
    & Stocks&\textbf{0.037$\pm$ 0.000}&\textbf{0.037$\pm$ 0.000}&0.039$\pm$ 0.000&0.038$\pm$ 0.000\\
    & Energy&0.251$\pm$ 0.011&\textbf{0.250$\pm$ 0.000}&0.338$\pm$ 0.010&0.277$\pm$ 0.001\\
    \bottomrule
    \end{tabular}
    \caption{TS generation results with generation length 24 for Sines, Stocks, and Energy datasets. 
    PaD-TS shows state-of-the-art performance in most cases. \textbf{Bold} font (lower score) indicates the best performance. Hyperparameters in Appendix \ref{app:exp}.}
    \label{Table:1}
\end{table*}

The final output can be obtained by adding all DiT blocks output from cross-dimension and temporal modules with a dense layer that converts to its original shape:
\begin{equation}
    x_{out} = (W_D^2 \sum_{i = 0}^{N} O_{D}^{i}+ b_D^2)  + (W_T^2 \sum_{i = 0}^{N} O_{T}^i + b_T^2)
\end{equation}

\section{Experiments}
In this section, we describe our experiment settings and evaluate the TS generation quality of PaD-TS across different domains and sequence lengths. The experiment results consist of quantitive and qualitative results in terms of \textit{individual} authenticity and \textit{population}-level property preservation. We also perform an ablation study to demonstrate the effectiveness of each proposed component and the effect of the hyperparameter $\alpha$.

\subsection{Experiment Settings}
We briefly discuss the datasets, baseline models, and evaluation methods. All experiments are run on a Rocky Linux server with AMD EPYC 7313 CPU, 128 GB of memory, and 2 Nvidia A40 GPUs. Additional model hyperparameters are provided in Appendix \ref{app:exp}.

\begin{figure}[t]
\centering
\includegraphics[width=1\columnwidth]{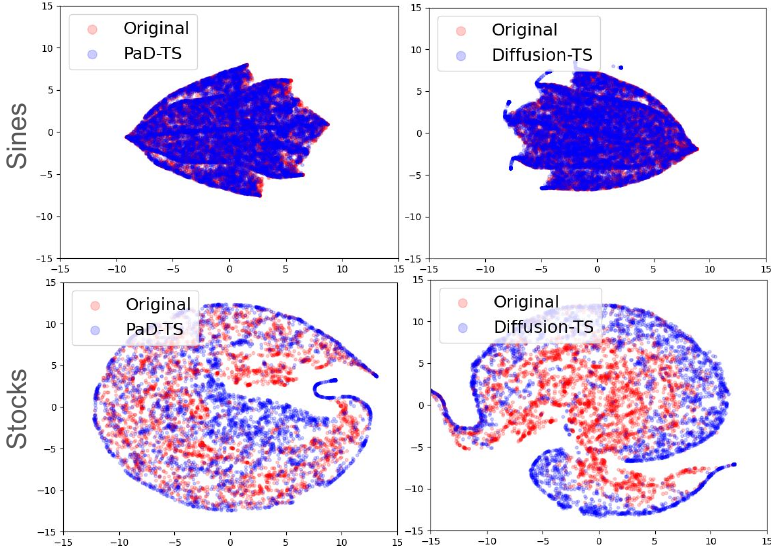} 
\caption{t-SNE plots on the cross-correlation values between original data (red dots) and synthetic data (blue dots) on the Sines and Stocks dataset. 
}
\label{figtsn-1}
\end{figure}

{\bf Datasets}: We use three major benchmark datasets, spanning domains such as physics, finance, and synthetic time series. (1) \textit{Sines} \cite{yoon2019time}: Synthetic sine wave time series data that can be sampled based on parameters. (2) \textit{Stocks} \cite{yoon2019time}: Google stocks history time series data includes 5 features such as Open, Close, Volume, etc. (3) \textit{Energy} \cite{candanedo2017data}: Home appliances' energy consumption time series data includes 28 features such as energy consumption, room temperatures, room humidity levels, and more. Additional Mujoco \cite{tunyasuvunakool2020dm_control} and fMRI \cite{smith2011network} dataset results are available in Appendix \ref{app:experiment}.

{\bf Baselines}: We carefully select three previous models that perform well and cover all three generative frameworks: (1) Diffusion-TS \cite{yuan2024diffusionts} is a DM with trends and Fourier-based layers. (2) TimeGAN \cite{yoon2019time} is a GAN-based model with RNN layers. (3) TimeVAE \cite{desai2021timevae} is a VAE-based model with convolution layers, trends blocks, and seasonal blocks.

\begin{figure}[t]
\centering
\includegraphics[width=0.965\columnwidth]{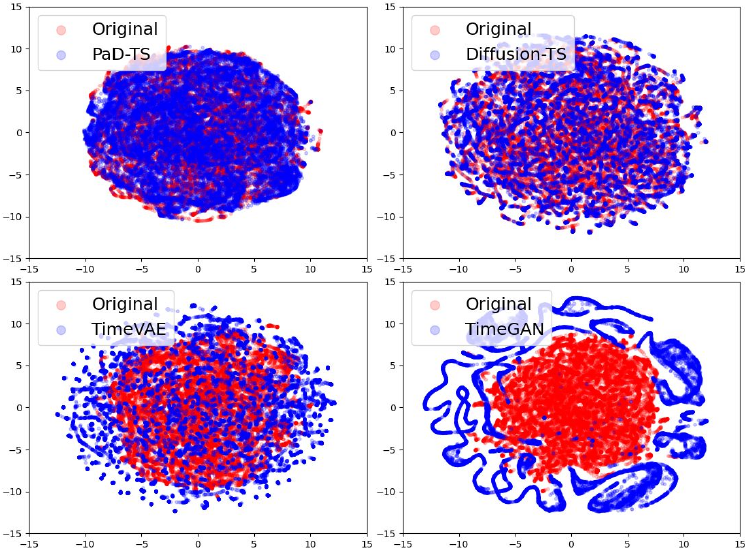} 
\caption{t-SNE plots on the cross-correlation values between original data (red dots) and synthetic data (blue dots) on the Energy dataset. 
}
\label{figtsn}
\end{figure}

\begin{table*}[hbt]
    
    \centering
    \begin{tabular}{llllll}
    \textbf{Metrics} & \textbf{Length} & \textbf{Pad-TS} & \textbf{Diffusion-TS} & \textbf{TimeGAN} & \textbf{TimeVAE}\\
    \toprule
    \multirow{3}{*}{VDS Score} 
    & 64 &  \textbf{0.0009} & 0.0043& 0.1688 & 0.0658\\
    &128 & \textbf{0.0005}& 0.0046& 0.1565 &0.0544\\
    &256 & \textbf{0.0008}& 0.0044& 0.2725 &0.0416\\
    \midrule
    \multirow{3}{*}{FDDS score} 
    & 64 &  \textbf{0.0087}& 0.0476& 0.8540 &0.2656\\
    &128 & \textbf{0.0009}& 0.0112& 0.9767 &0.1120\\
    &256 & \textbf{0.0010}& 0.0038&3.0019&0.0424\\
    \midrule
    \multirow{3}{*}{DA} 
    & 64 &  \textbf{0.023} $\pm$ \textbf{0.009} & 0.094 $\pm$ 0.009 & $0.437 \pm 0.062$ &$0.499 \pm 0.000$\\
    &128 & \textbf{0.050} $\pm$ \textbf{0.080} & 0.165 $\pm$ 0.067 & $0.399 \pm 0.268$ &$0.499 \pm 0.000$\\
    &256 & \textbf{0.138} $\pm$ \textbf{0.174} & 0.393 $\pm$ 0.009 & $0.499 \pm 0.000$ &$0.492 \pm 0.001$\\
    \midrule
    \multirow{3}{*}{Predictive Score} 
    & 64 &  \textbf{0.248} $\pm$ \textbf{0.000} & 0.249 $\pm$ 0.000 & $0.301 \pm 0.007$ & $0.290 \pm 0.001$\\
    &128 & \textbf{0.247} $\pm$ \textbf{0.003} & 0.248 $\pm$ 0.001 & $0.316 \pm 0.008$ & $0.290 \pm 0.000$\\
    &256 & \textbf{0.244} $\pm$ \textbf{0.001} & 0.250 $\pm$ 0.002 & $0.285 \pm 0.006$ & $0.266\pm 0.001$\\
    \bottomrule
    \end{tabular}
    \caption{Long TS Generation Results on Energy dataset. \textbf{Bold} font (lower score) indicates the best performance.}
    \label{Table:long-seq}
\end{table*}

{\bf Evaluation metrics}: We use the following metrics to evaluate the TS generation quality: (1) VDS score, (2) FDDS score, (3) DA score, and (4) predictive score. The first three metrics are introduced in Section \ref{PS}, where VDS and FDDS scores measure the population-level distribution shift of generated TS in terms of value and functional dependency. We use CC as an example of functional dependency. DA score \cite{yoon2019time} measures the individual-level authenticity. In addition, we are interested in evaluating how the generated data can be used in downstream tasks such as TS prediction. Hence, our last metric is the predictive score \cite{yoon2019time}, which is the mean absolute error score of the TS prediction result where the post-hoc RNN model is trained using synthetic TS data and evaluated on real TS data. Due to the unstable nature of the DA score and predictive score, we repeat the evaluation for 5 iterations and report the mean and standard deviation for robust results.
We additionally include feature and distance-based metrics summarized in TSGBench \cite{ang2023tsgbench}: Marginal Distribution Difference (MDD), AutoCorrelation Difference (ACD), Skewness Difference (SD), Kurtosis Difference (KD), Euclidean Distance (ED), and Dynamic Time Warping (DTW).

\begin{table}
    \centering
    \begin{tabular}{lll}
    \textbf{Metrics} & \textbf{PaD-TS} & \textbf{Diffusion-TS}\\
    \toprule
    MDD  & 0.609 & \textbf{0.573}  \\
    ACD & \textbf{0.061} & 0.200\\
    SD & 0.027 & \textbf{0.025}\\
    KD & \textbf{0.032}& 0.049\\
    ED & \textbf{0.645} & 0.658\\
    DTW & \textbf{1.674} & 1.718\\
    \bottomrule
    \end{tabular}
    \caption{Feature and distance-based measures comparison between Diffusion-TS and PaD-Ts on Sines dataset. {Bold} font (lower score) indicates the best performance.}
    \label{Table:sineTSG}
\end{table}

\subsection{Comparison with Baselines} 
In Table \ref{Table:1}, we present the result for a benchmark setting that is commonly used in state-of-the-art TS generation models \cite{yuan2024diffusionts,yoon2019time}: TS generation with sequence length 24. The result shows that PaD-TS consistently outperforms previous methods in terms of population-level property preservation (i.e., VDS and FDDS). Averaging across all three datasets, PaD-TS improves the FDDS score by 5.9x and the VDS score by 5.7x compared to the previous state-of-the-art model, Diffusion-TS, while maintaining comparable performance in individual-level authenticity. In Table \ref{Table:sineTSG}, Table \ref{Table:stockTSG} and Table \ref{Table:energyTSG}, we present the feature and distance-based metrics results. PaD-TS achieved comparable or better performance across Sines, Stocks, and Energy datasets.

To better understand the performance of population-level preservation in different models, we visualize the t-distributed stochastic neighbor embedding (t-SNE) \cite{van2008visualizing} on the functional dependency cross-correlation values obtained from the Sines, Stocks, and Energy datasets and their corresponding synthetic data in a lower-dimensional space. 
As we can see in Figure \ref{figtsn-1} and \ref{figtsn}, the t-SNE plot produced by PaD-TS shows the best alignment with that of the original data (red dots). This is consistent with Table \ref{Table:1}, where the FDDS score of PaD-TS is significantly lower than the baselines. In addition to t-SNE, we include more visualizations of the global value distribution in the 
Appendix \ref{app:figs}. 
The results indicate that PaD-TS is also the most effective in preserving value distributions.
\begin{table}
    \centering
    \begin{tabular}{lll}
    \textbf{Metrics} & \textbf{PaD-TS} & \textbf{Diffusion-TS}\\
    \toprule
    MDD  & \textbf{0.379} & 0.440  \\
    ACD & 0.111 & \textbf{0.028}\\
    SD & \textbf{0.375} & 0.471\\
    KD & 4.290& \textbf{2.207}\\
    ED & 1.135 & \textbf{1.093}\\
    DTW & 2.937 & \textbf{2.829}\\
    \bottomrule
    \end{tabular}
    \caption{Feature and distance-based measures comparison between Diffusion-TS and PaD-Ts on Stocks dataset. {Bold} font (lower score) indicates the best performance.}
    \label{Table:stockTSG}
\end{table}

\textbf{Long sequence generation.} We further challenge PaD-TS in TS generation with longer sequence lengths (64, 128, and 256) on the high-dimensional Energy dataset. 
The results in Table \ref{Table:long-seq}
show that PaD-TS has a dominating performance compared to baselines in all 4 metrics. PaD-TS not only improved the FDDS score by 6.1x on average but also made a significant improvement in the DA score by 3.4x on average.

\textbf{Time Complexity Comparison.}
PaD-TS requires slightly longer training time compared to existing DMs but remains reasonable. Two primary factors contribute to the extended training time: 1) the additional loss term, $L_{pop}$, which computes the pairwise distribution distance, and 2) the SSS, which necessitates additional iterations relative to the standard sampling strategy. In table \ref{Table:time}, we present the training time required between Diffusion-TS and PaD-TS for selected datasets.

\begin{table}
    \centering
    \begin{tabular}{lll}
    \textbf{Metrics} & \textbf{PaD-TS} & \textbf{Diffusion-TS}\\
    \toprule
    MDD  & 0.221 & \textbf{0.200}  \\
    ACD & \textbf{0.055} & 0.141\\
    SD & \textbf{0.124} & 0.174\\
    KD & \textbf{1.037}& 1.387\\
    ED & \textbf{1.030} & 1.032\\
    DTW & \textbf{6.395} & 6.439\\
    \bottomrule
    \end{tabular}
    \caption{Feature and distance-based measures comparison between Diffusion-TS and PaD-Ts on Energy dataset. {Bold} font (lower score) indicates the best performance.}
    \label{Table:energyTSG}
\end{table}

\begin{table}
    \centering
    \begin{tabular}{lll}
    \textbf{Dataset} & \textbf{PaD-TS} & \textbf{Diffusion-TS}\\
    \toprule
    Sines & 77min& 17min\\
    Stocks & 75min & 15min\\
    Energy & 117min & 60min\\
    \bottomrule
    \end{tabular}
    \caption{Training time comparison between PaD-TS and Diffusion-TS.}
    \label{Table:time}
\end{table}

\begin{figure}[ht]
\centering
\includegraphics[width=0.95\columnwidth]{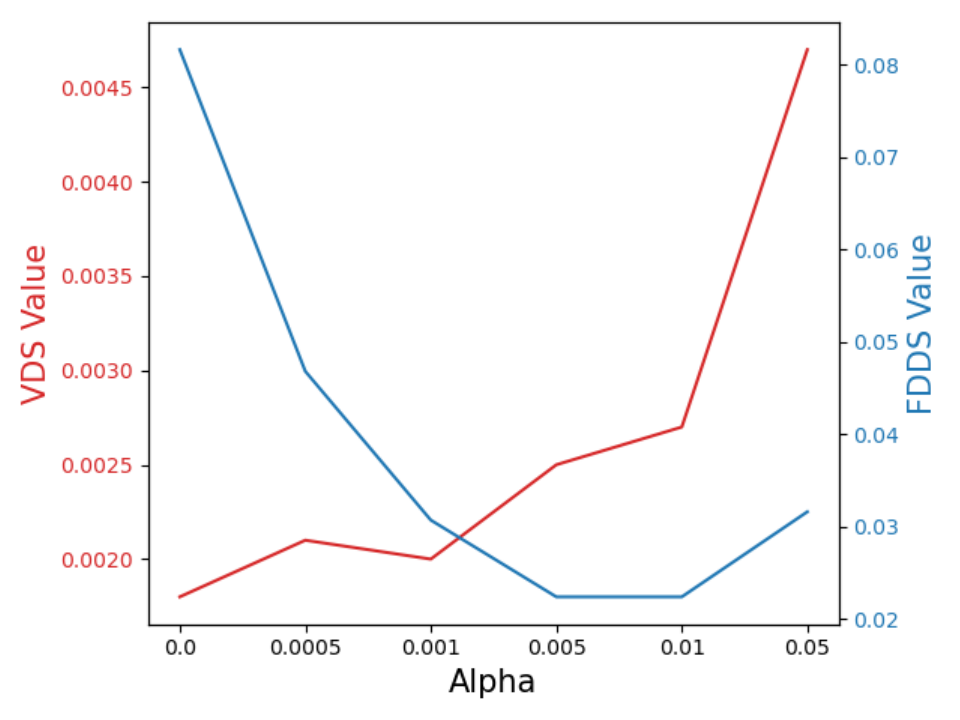} 
\caption{Ablation study on $\alpha$ and Energy dataset. The blue and red curves resp. depict the FDDS and VDS scores.}
\label{Fig:ablation}
\end{figure}

\subsection{Ablation Study}
To further understand our model, we conduct two ablation studies to evaluate: (1) the effectiveness of each model component in terms of population-level property (i.e., CC) preservation, and (2) the effect of the population aware training objective hyperparameter $\alpha$.

(1) In the first ablation study, we train PaD-TS variants by taking out one of the four components of the full PaD-TS as follows: (1) without (w/o) temporal channel, (2) w/o dimension channel, (3) w/o PAT objective, and (4) w/o SSS strategy. In Table \ref{Table:ablation}, results indicate that the SSS strategy and temporal channel are the most useful components, while the PAT training objective and the dimension channel are less effective but still crucial. By combining all four components, the full PaD-TS variant shows the best performance.

(2) In the second ablation study, we train different PaD-TS models with different $\alpha$ values ranging from 0 to 0.05. Intuitively, a larger $\alpha$ indicates more weight to the CC distribution loss $L_{pop}$ and less weight to the original loss $L_0$. 
In Figure \ref{Fig:ablation}, results show that when $\alpha$ increases, there is a general trend of increasing (worse) VDS score and decreasing (better) FDDS score. Once $\alpha$ goes too large ($\alpha=0.05$), the entire training collapses with large VDS and FDDS scores. 
\begin{table}
    \centering
    \begin{tabular}{lllll}
    \textbf{Metrics} & \textbf{Model} & \textbf{Sines} & \textbf{Stocks} & \textbf{Energy}\\
    \toprule
    \multirow{4}{*}{FDDS} 
    & PaD-TS & \textbf{0.0003} & \textbf{0.0588} & \textbf{0.0442}\\
    & w/o Temporal & 0.0005  & 	1.8838 & 0.5254 \\
    & w/o Dimension & 0.0087 & 0.0868 & 0.0533\\
    & w/o PAT &  0.0007 & 0.2459 & 0.0816\\
    & w/o SSS & 0.0286 & 0.0965 & 0.3626 \\
    \bottomrule
    \end{tabular}
    \caption{Ablation study for the effectiveness of PaD-TS components. \textbf{Bold} font indicates the best performance.}
    \label{Table:ablation}
\end{table}

\section{Conclusion}
We study the TS generation problem with a focus on the preservation of TS population-level property. Towards this goal, our core contribution is PaD-TS, a novel DM that is equipped with a new population-aware training process, and a new dual-channel encoder model architecture. Our extensive experimental results show that PaD-TS achieves state-of-the-art performance both qualitatively and quantitatively in all three benchmark datasets over the two population-level authenticity metrics. Our ablation study also shows the effectiveness of each new component in PaD-TS. In the future, we would like to further enhance PaD-TS with the ability to do conditional generation (e.g., constrained by certain trend information), and apply PaD-TS to downstream TS-related tasks in low-resource domains, especially where generation bias could lead to critical issues.

\section*{Acknowledgements}
This work was supported in part by Generated Health and The William \& Mary Applied Research \& Innovation Initiative Exploratory Award. 
\bibliography{aaai25}

\appendix
\newpage
\onecolumn

\section{Cross-Correlation} \label{CC_detail}
\bigskip

In this section, we briefly review the definition of CC. 
CC and its normalized form Pearson correlation are popular similarity measurements in TS and signal processing tasks \cite{liao2020conditional,yuan2024diffusionts}. Given two univariate TS $X_t$ and $Y_t$ where $t$ stands for time, the general CC function $R_{X_t Y_t}$ can be formulated as:
\begin{equation}
    R_{X_t Y_t}(\tau) = \mathbb{E}[X_{t-\tau}Y_t]
\end{equation}
where $\tau$ stands for optional lags. Throughout the experiment, we set $\tau = 0$ and apply it for later derivations.

By further normalizing $X_t$ and $Y_t$ with their means $\mu_X$ and $\mu_Y$, we will obtain cross-covariance $K_{XY}$ between them:
\begin{equation}
    K_{X_t Y_t} = \mathbb{E}[(X_{t} - \mu_X)(Y_t - \mu_Y)]
\end{equation}

Finally, we can normalize $K_{XY}$ with variance measures $\sigma_X^2$ and $\sigma_Y^2$ to obtain the Pearson correlation coefficient $\rho$, which is more interpretable and ranges from  [-1,1] with the following:
\begin{equation}
    \begin{aligned}
         \rho_{X_t Y_t} &= \frac{\mathbb{E}[(X_{t} - \mu_X)(Y_t - \mu_Y)]}{\sqrt{\sigma_X^2} \sqrt{\sigma_Y^2}} \\
                   &= \frac{\mathbb{E}[X_t Y_t] - \mathbb{E}[X_t]\mathbb{E}[Y_t]}{\sqrt{\mathbb{E}[X^2_t]- (\mathbb{E}[X_t])^2} \sqrt{\mathbb{E}[Y^2_t]- (\mathbb{E}[Y_t])^2}}
    \end{aligned}
\end{equation}

\newpage
\section{Diffusion Transformer (DiT) Blocks} \label{app:DiT}
\bigskip

\begin{figure}[ht]
\centering
\includegraphics[width=0.9\columnwidth]{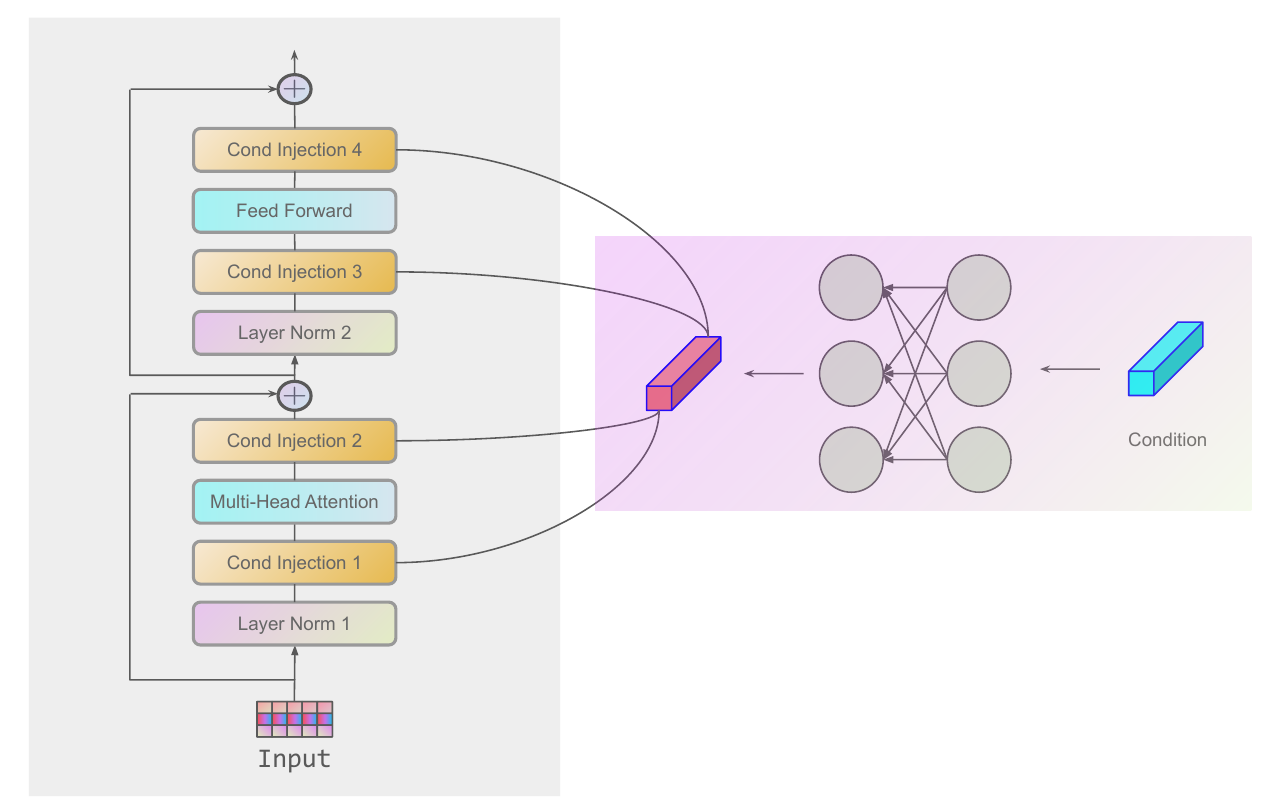} %
\caption{DiT block architecture.}
\label{Fig:DiT}
\end{figure}

In this section, we briefly review the architecture of DiT blocks \cite{peebles2023scalable}. 
DiT blocks contain more parameters and offer high throughput during training. The experiments demonstrate promising generation quality and strong scalability.
As illustrated in Figure \ref{Fig:DiT}, the DiT block architecture closely resembles that of a standard transformer encoder. For the conditional input 
$c$, DiT divides the hidden state into 6 chunks and gradually introduces them using an adaLN-Zero design. Condition injection layers 1 and 3 incorporate two chunks of the conditional hidden features each, while layers 2 and 4 incorporate one chunk each. Unlike vanilla transformer encoder-based models, DiT can generate samples based on conditional information.

\newpage
\section{Additional Experiment Details} \label{app:exp}
\bigskip
In this section, we discuss the code implementation details and the hyperparameters we explored throughout the experiment. 

Our code is based on the improved DDPM \cite{nichol2021improved} as it is more formally implemented with additional features such as different sampling strategies, models towards different targets ( $x^0$, $x^t$ and $\epsilon^t$), etc. For the model architecture, we adopted some implementation from diffusion transformer \cite{peebles2023scalable} and inverted transformer \cite{liu2024itransformer}. Finally, we modify code from TimeGAN \cite{yoon2019time} and Diffusion-TS \cite{yuan2024diffusionts} for data pre-processing and evaluation. 

The hyperparameters of a DM usually come from two sources: (1) the general DM pipeline and (2) the model architecture.

(1) For the general DM pipeline, we use a cosine noise scheduler and a model toward input without noise $x^0$ throughout our experiment. We additionally tuned the following hyperparameters: diffusion steps from 100 to 700, batch sizes from 32 to 128, and normalization strategies (min-max or -1 to 1). 

(2) For the model architecture, we use an AdamW optimizer \cite{loshchilov2017decoupled} (with learning rate $= 0.0001$), the proposed PaD-TS architecture, and 4 attention heads in all transformer-related blocks. We additionally tuned our model hyperparameters: $\alpha$ from 0 to 0.05, hidden dimension from 32 to 256, number of encoders from 1 to 2, and number of DiT blocks from 2 to 4.  We found the following set of best-working hyperparameters listed in Table \ref{Table:parameter}:
\bigskip
\begin{table*}[th]
    \centering
    \begin{tabular}{llll}
    \textbf{Parameter} & Sines & Stocks & Energy\\
    \toprule
    Target & $x^0$& $x^0$& $x^0$\\
    Noise Scheduler & Cosine & Cosine & Cosine\\
    Diffusion Step & 250& 250& 500 \\
    Batch Size& 64& 64& 64 \\ 
    Normalization& -1 to 1& -1 to 1& -1 to 1\\
    \midrule
    Optimizer & AdamW& AdamW& AdamW\\
    Num of Heads & 4& 4& 4\\
     $\alpha$ & 0.0005& 0.0008 & 0.0005\\
    Hidden Dim & 128&128 & 256\\
    Num of Enc & 1 & 1 & 1\\ 
    Num of DiTs& 3 & 3 & 3\\
    \bottomrule
    \end{tabular}
    \caption{List of DM and model-related parameters with generation length 24 for Sines, Stocks, and Energy dataset. DM parameters are listed in the first half, and model-related parameters are listed in the second half of the Table.}
    \label{Table:parameter}
\end{table*}

\newpage
\section{Additional Figures} \label{app:figs}
\bigskip

\begin{figure}[th]
\centering
\includegraphics[width=\columnwidth]{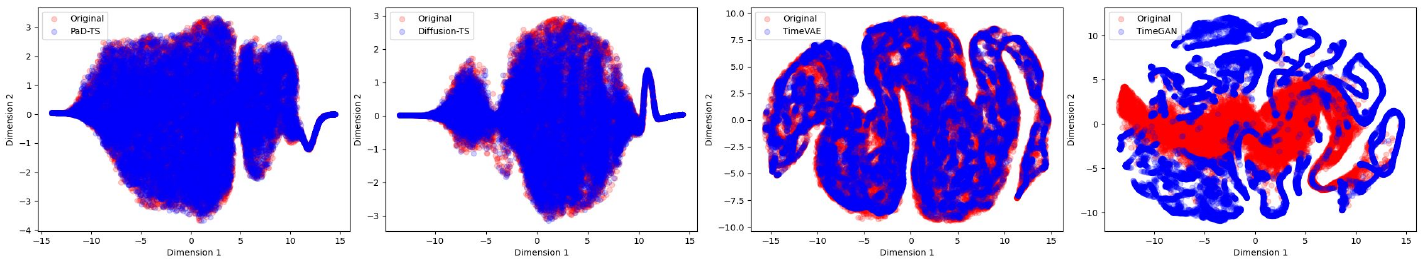} 
\caption{t-SNE plots on the average values for each dimension between original data (red dots) and synthetic data (blue dots) on the Energy dataset.}
\label{Fig:TSN_VD}
\end{figure}

This section provides additional figures that compare the value distributions between the original and synthetic data. Figure \ref{figtsn} displays a t-SNE plot of the CC values for both the original and synthetic data on the Energy dataset, highlighting the performance in preserving the CC distribution. Following Diffusion-TS \cite{yuan2024diffusionts}, we use t-SNE and data distribution plots to compare how well different methods maintain the value distribution. 
As shown in Figure \ref{Fig:TSN_VD}, the results indicate that PaD-TS is more closely aligned with the original data set (red dots).

\bigskip

\begin{figure}[th]
\centering
\includegraphics[width=\columnwidth]{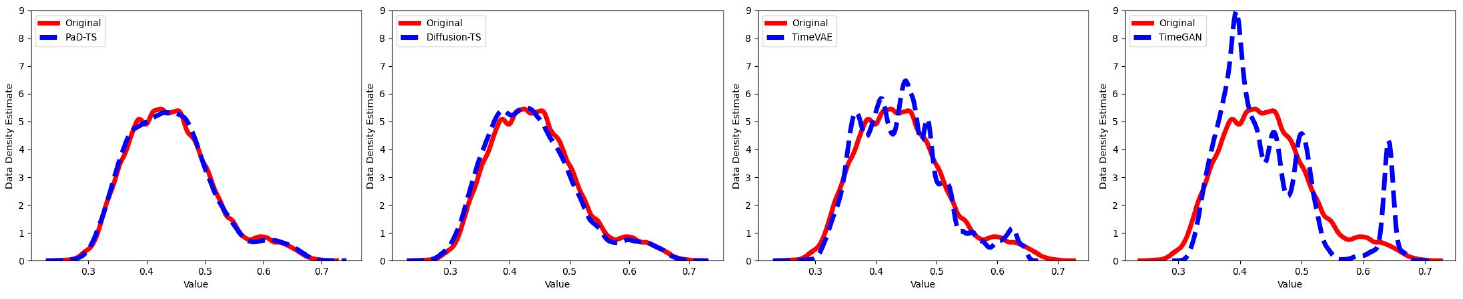} 
\caption{Value distribution plots on the average values for each dimension between original data (red line) and synthetic data (blue line) on the Energy dataset.}
\label{Fig:VD}
\end{figure}

For the value distribution plot Figure \ref{Fig:VD}, we also see the PaD-TS best aligns with the original dataset (red line), which is consistent with results in Figure \ref{Fig:TSN_VD} and Table \ref{Table:1}.

\bigskip

\newpage
\section{Additional Experiment} \label{app:experiment}
\bigskip
This section compares the PaD-TS performance with Diffusion-TS on the Mujoco and the fMRI datasets. In Table \ref{Table:mujoco}, PaD-TS achieves performance on par with Diff-TS in terms of discriminative accuracy and predictive score, along with improvements in FDDS and VDS.

\begin{table}[H]
    \centering
    \begin{tabular}{llll}
    \textbf{Metrics} & \textbf{Dataset} & \textbf{PaD-TS} & \textbf{Diffusion-TS} \\
    \toprule
    \multirow{2}{*}{VDS} 
    & fMRI &\textbf{0.0008} &0.0010\\
    & Mujoco&\textbf{0.0009} &0.0014\\
    \midrule
    \multirow{2}{*}{FDDS} 
    & fMRI &\textbf{0.0034}&0.0046\\
    & Mujoco&\textbf{0.0092}&0.0164\\
    \midrule
    \multirow{2}{*}{DA} 
    & fMRI &\textbf{0.153$\pm$ 0.032}&0.164$\pm$ 0.015\\
    & Mujoco&\textbf{0.016 $\pm$ 0.005}&0.018$\pm$ 0.009\\
    \midrule
    \multirow{2}{*}{Predictive score} 
    & fMRI &\textbf{0.100$\pm$ 0.000}&\textbf{0.100$\pm$ 0.000}\\
    & Mujoco&\textbf{0.008$\pm$ 0.002}&\textbf{0.007$\pm$ 0.001}\\
    \bottomrule
    \end{tabular}
    \caption{TS generation results with generation length 24 for Mujoco and fMRI datasets. \textbf{Bold} font (lower score) indicates the best performance. Hyperparameters are listed in the code repo.}
    \label{Table:mujoco}
\end{table}

\end{document}